\documentclass[11pt,a4paper,final]{article}
\usepackage[left=1in,right=1in,top=1in,bottom=1in]{geometry}
\usepackage{palatino} 
\usepackage{authblk}
\PassOptionsToPackage{sort&compress}{natbib}
\usepackage[]{natbib}

\usepackage{booktabs} 
\usepackage[ruled]{algorithm2e} 
\usepackage{tikz}
\usetikzlibrary{arrows}
\usetikzlibrary{shapes.multipart}
\usepackage{longtable}
\usepackage{caption}

\SetAlFnt{\small}
\SetAlCapFnt{\small}
\SetAlCapNameFnt{\small}
\SetAlCapHSkip{0pt}
\IncMargin{-\parindent}
\usepackage[all]{nowidow}

\usepackage[font=small,labelfont=bf]{caption}
\usepackage[utf8]{inputenc} 
\usepackage[T1]{fontenc}    
\usepackage{hyperref}       
\hypersetup{
    colorlinks=false,
    linkcolor=blue,
    filecolor=magenta,      
    urlcolor=cyan,
    pdfpagemode=FullScreen,
    }
\usepackage{url}            
\usepackage{booktabs}       
\usepackage{amsfonts}       
\usepackage{nicefrac}       
\usepackage{microtype}      
\usepackage{xcolor}         
\usepackage{amsmath}
\usepackage{mathtools}
\usepackage{amsthm}
\usepackage{amsmath}
\usepackage{amssymb}
\usepackage{bbm}
\usepackage{tikz} 
\usepackage{wrapfig}
\usepackage{floatrow}
\usepackage{enumitem}

\usepackage[suppress]{color-edits}
\usepackage{caption}
\usepackage{subcaption}

\usepackage{xcolor}

\title{Revisiting Rogers' Paradox in the Context of Human-AI Interaction}

\begin{document}

\author[1]{Katherine M. Collins}
\author[2,3]{Umang Bhatt}
\author[2]{Ilia Sucholutsky}
\affil[1]{University of Cambridge}
\affil[2]{New York University}
\affil[3]{The Alan Turing Institute}
\date{}

\maketitle

\begin{abstract}

Humans learn about the world, and how to act in the world, in many ways: from individually conducting experiments to observing and reproducing others' behavior. Different learning strategies come with different costs and likelihoods of successfully learning more about the world. The choice that any one individual makes of how to learn can have an impact on the collective understanding of a whole population if people learn from each other. Alan Rogers developed simulations of a population of agents to study these network phenomena where agents could individually or socially learn amidst a dynamic, uncertain world -- and uncovered a confusing result: the availability of cheap social learning yielded no benefit to population fitness over individual learning. This paradox (Rogers' Paradox) spawned decades of work trying to understand this equilibrium and uncover factors that foster the relative benefit of social learning that centuries of human behavior suggest exists. But what happens in such network models now that humans can socially learn from AI systems that are themselves socially learning from us? We revisit Rogers' Paradox in the context of human-AI interaction and extend the simulations introduced in Rogers' Paradox to examine a simplified network of humans and AI systems learning together about an uncertain world. We propose and examine the impact of several learning strategies on the quality of the equilibrium of a society's ``collective world model''. We consider strategies that can be undertaken by various stakeholders involved in a single human-AI interaction: the human, the AI model builder, and the society or regulators around the interaction. We then extend the environment model to consider possible negative feedback loops that may arise from humans learning socially from AI: that learning from the AI may impact our own ability to learn about the world. We close with several open directions into studying networks of human and AI systems that can be explored in enriched versions of our simulation framework.  

\end{abstract}

\section{Introduction}

For centuries, humans have learned about the world in different ways: from each other, and from individually conducting experiments and exploring the world around us. From young children ~\citep{gopnik1996scientist} to pioneers like Isaac Newton, Marie Curie and Archimedes, humans have long engaged with the world and each other in many ways like scientists -- tinkering with our models of the world~\citep{bramley2023local, rule2020child} and sharing this knowledge to impact society's collective understanding of the world. 

But conducting experiments yourself -- whether exploring out in the world, or thinking really hard to process what you have already observed -- often comes with costs. It takes time and energy to think and to explore, and we have fundamental constraints on such resources~\citep{lieder2020resource, griffiths2020understanding}. Sometimes, it is easier to just build on the behavior or insights from another person; to update your understanding of the world based on what someone else has done~\citep{rendell2011cognitive}. Humans do this all the time -- from reading books written by great thinkers, to studying new sports moves by observing athletes on TV, to watching a cooking video, to learning how to make the local food of a distant land. To make sure that each of us has a deeper understanding of our ever-changing world, we need to adequately modulate the relative costs of social versus individual learning to make good decisions and plan. 

The question of the relative advantage of social versus individual learning among humans was thrown for a loop when Alan Rogers uncovered a seeming paradox: the availability of cheap social learning does not increase the relative fitness of a population compared to a population consisting entirely of individual learners~\citep{rogers1988does}. This has led to many follow-ups exploring what strategies any single learner can employ to mitigate this challenge, for example, shifting to a ``critical'' social learning approach where you toggle between individual and social learning depending on your estimate of the expected utility~\citep{enquist2007critical} or otherwise evaluating who to learn from based on other characteristics of the other agent, such as their age or apparent prestige, or other characteristics of the information, e.g., how long it has been around~\citep{lew2023peer, deffner2022does, deffner2020dynamic, miu2020flexible}.

While there remain many rich questions on how humans engage with other \textit{humans} \citep{wu2024group}; here, we advocate for reconsidering Rogers' Paradox in the context of today's more powerful and, at the surface, human-compatible AI systems, specifically systems that can engage with language. To date, we can think of these systems as having ``socially learned'' from us: they have been trained on effectively all of what we humans have written on the web~\citep{andreas2022language}. To an extent, any ``world model'' or other ``understanding of the world'' that may or may not have been implicitly learned within one of these models~\citep{yildirim2024task, hao2023reasoning, ivanova2024elements, mitchell2023ai, liemergent, mitchell2023debate, 
gurnee2023language, vafa2024world} could then arguably have been socially learned from our actions and experiments in the world, that we wrote about in language\footnote{Here, we define ``world model'' as queryable understanding of the world. We refer to AI systems learning a collective world model abstractly (as if there was a single, global ``ideal'' model). We are not claiming that current large language models have or have not learned such a world model, nor whether such a global model is feasible.}. The fluency with which these systems can engage with natural language, and therefore us, raises the prospect that now and in the ensuing years, the directionality of learning may flip: humans may increasingly socially learn from these systems ~\citep{collins2024building, brinkmann2023machine}. While work has already begun to explore what may unfold if AI systems learn from their own output, or the output of other AI systems~\citep{shumailov2024ai} --- here, we focus on possible trajectories from humans and AI systems learning together.


There is an exciting and pressing need to understand human-AI interactions, and how these interactions may impact our understanding of the world around us. Various works have studied what can be done to make a single ``dyadic'' human-AI interaction ``go right'' (that is, a human interacting with an AI system)~\cite{lai2023towards,gabriel2024ethics}. While there is much value in understanding interactions at the scale of one unit -- and there is much to still be uncovered in how these interactions can add value beyond what any one agent can do~\citep{vaccaro2024combinations} -- in this work, we take initial steps to understand a \textbf{network of agents}, wherein many humans interact with a single AI system (where the AI system also learns from people), and characterize interconnected ripple effects on a \textit{society} of such learning agents. 

To take initial steps to probe possible network effects, we extend the population model introduced in Rogers Paradox with a simulated AI agent: an agent that learns from all other agents. We emphasize that our work then focuses on possible learning \textit{equilibria} induced by a society of different agents, with distinct learning strategies and costs, in a way that could mimic an abstract network model of humans engaging with an AI system. We emphasize, however, that our ``AI system'' and simulated ``human-AI interactions'' are highly simplistic, abstract models of agents and their interaction; part of the power, as in much complex systems research, comes from the simulation of many such interactions with simple components ~\citep{krakauer2023unifying,mitchell2006complex}. Rogers' Paradox and the follow-up works spawned to understand and mitigate Rogers' Paradox have contributed to a richer conceptual understanding of collective behavior and inspired new empirical work, despite -- and sometimes because of -- necessarily simplistic simulations~\citep{wu2024group,perry2022makes,rendell2010copy, rendell2011cognitive}. Here, we too hope that revisiting Rogers' Paradox in the context of human-AI interaction will inspire new ways of thinking about and studying possible network propagation effects of AI systems among previously well-characterized networks of human individual and social learners where the world is fundamentally uncertain and dynamic. 




To that end, this work is structured as follows. We first (Section~\ref{sec2}) review Rogers' Paradox and extend the network model to consider an AI system (with different costs) that learns from the entire population at once is introduced into the network. We then (Section~\ref{sec3}) consider possible strategies that can be enacted by different actors involved in the interaction and network of interactions: the human, the AI model builders, the interface designers around the moment of interaction, and policymakers writ large. We then turn to a different model of network dynamics (Section~\ref{negative-feedback}) wherein interacting with an AI system impacts an agent's ability to individually learn. We close in (Section~\ref{looking-ahead}) by discussing other questions raised by our ``AI Rogers' Paradox'' simulations, including ways to enrich the realism of the simulation and possible benefits of AI systems in the context of social learning (e.g., helping humans learn better from each other). 

\begin{figure}
    \centering
    \includegraphics[width=0.49\linewidth]{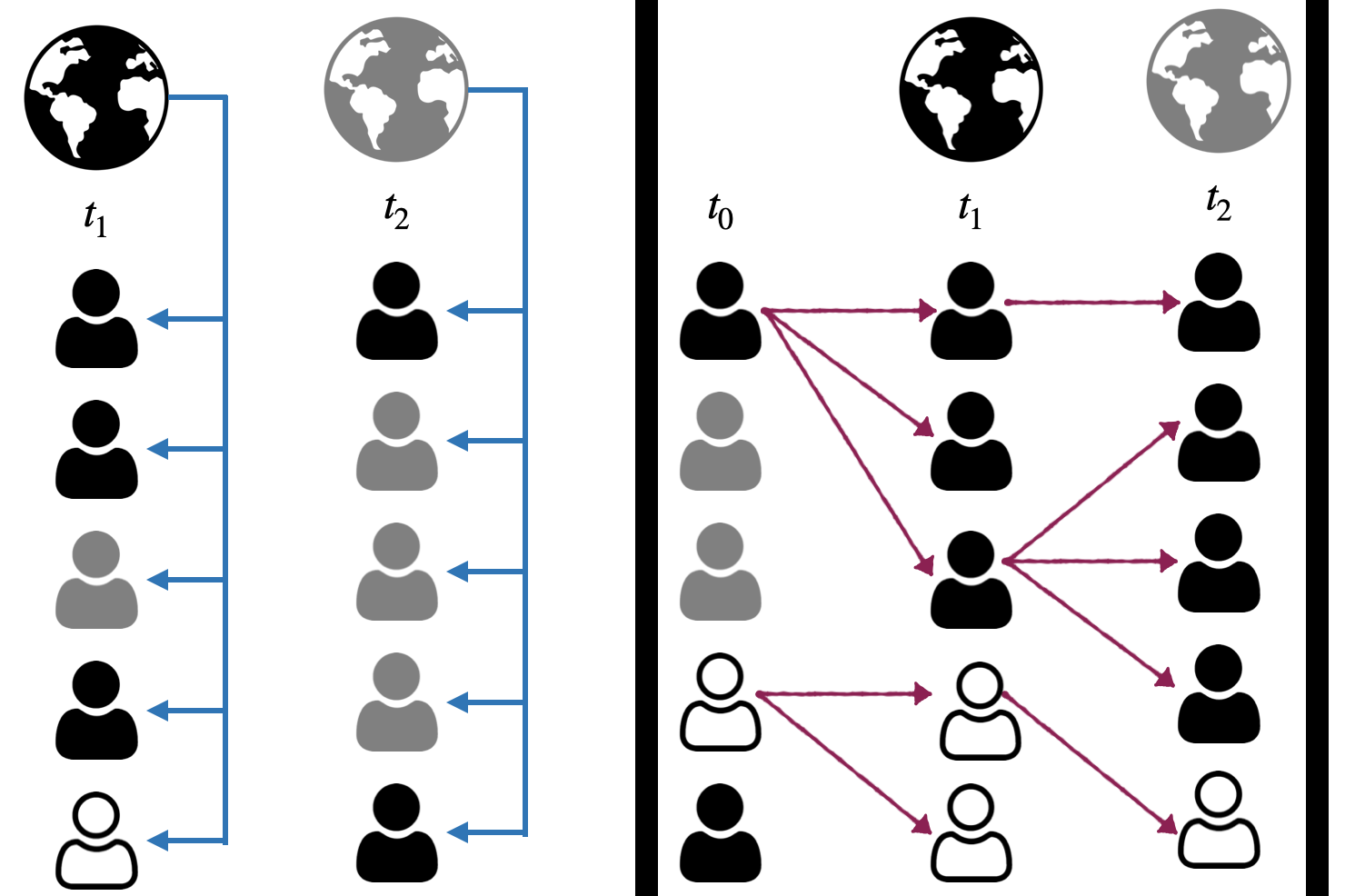}
    \includegraphics[width=0.49\linewidth]{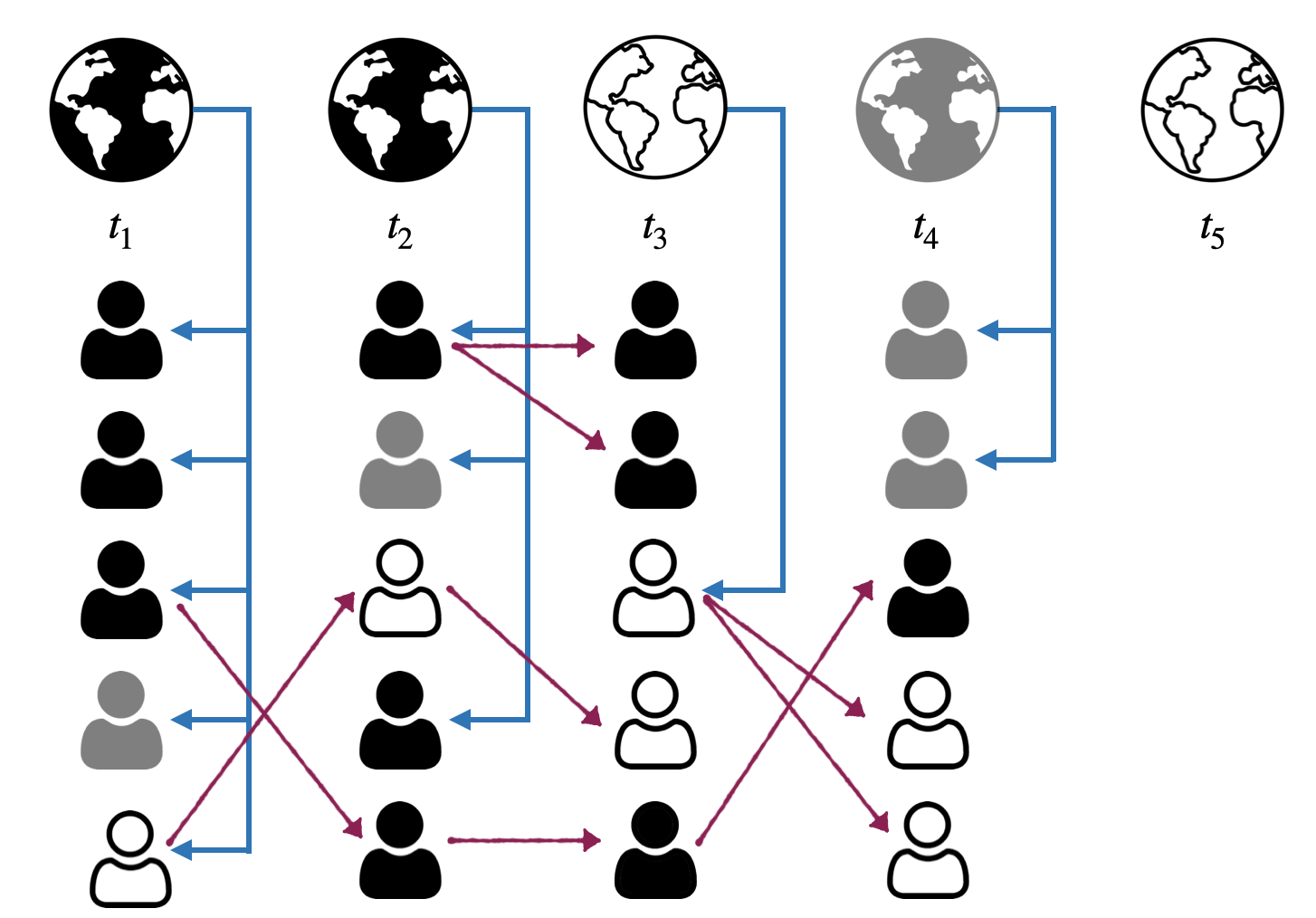}
    \caption{\textbf{Traditional Rogers' Paradox.} In the leftmost panel, we depict individual learning (blue arrows). In the middle panel, we depict social learning (purple arrows). Note that social learning is delayed by one timestep. In the rightmost panel, we depict an example of humans oscillating between individual and social learning. Agents are colored by the behavior they adopt in the current timestep. Agents are considered ``adapted'' if their behavior matches the current state of the world, and have an increased probability of surviving to the next timestep. }
    \label{fig:schematic}
\end{figure}

\begin{figure}
    \centering
    \includegraphics[width=0.85\linewidth]{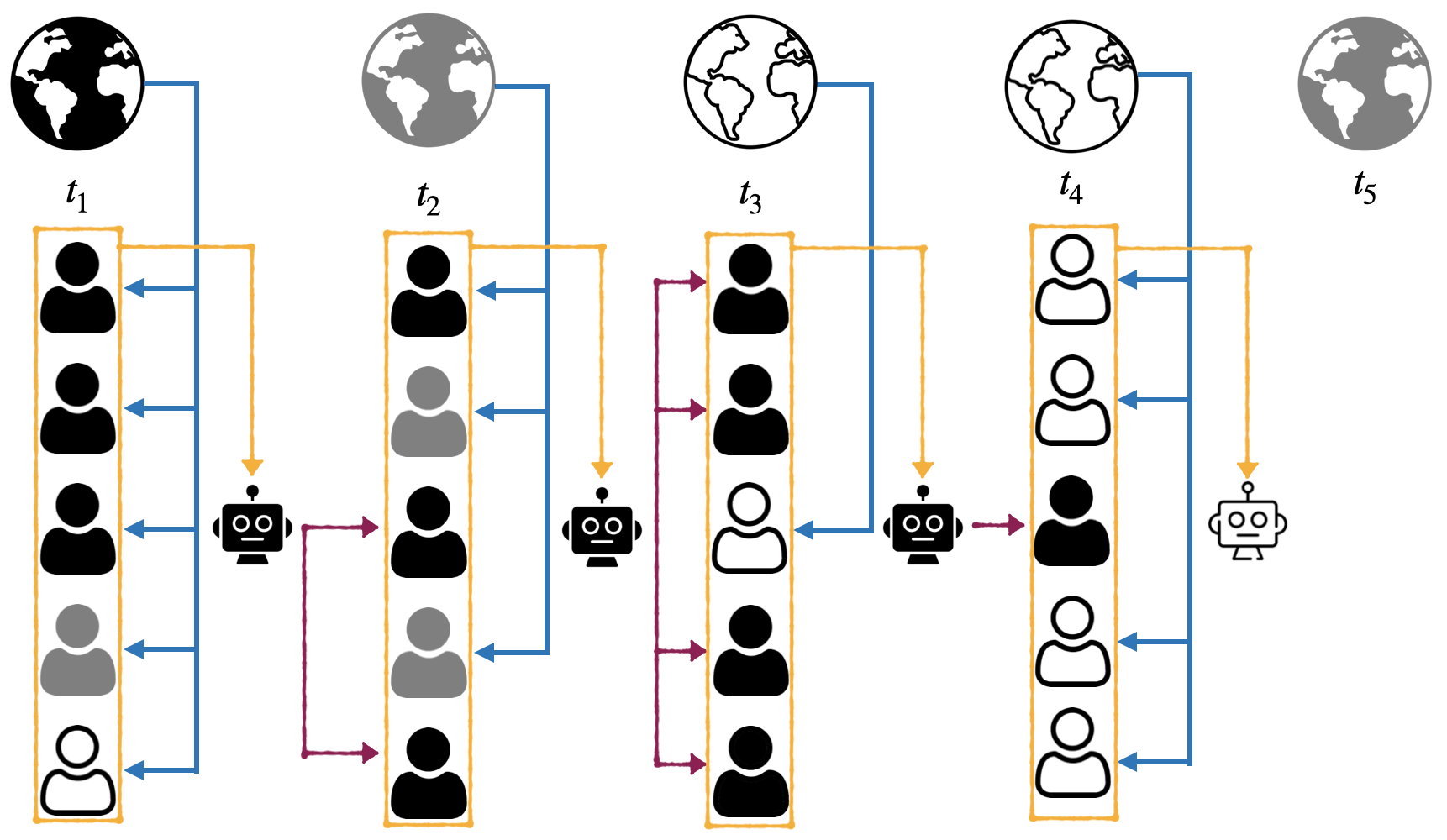}
    \caption{\textbf{AI Rogers' Paradox.} At each time step, humans can perform individual learning or learn from the AI system, which reverts to the population mean of the previous time step.}
    \label{fig:schematic}
\end{figure}

\section{Integrating Human-AI Interaction in Rogers' Paradox}
\label{sec2}


We first introduce Rogers' Paradox and the simulation environment introduced in ~\citep{rogers1988does}, see Figure~\ref{fig:schematic} for a visual summary. Rogers' Paradox considers the case where agents in a changing environment try to adapt their individual (simplified) ``world models'' by either learning about the environment individually or by learning socially from a number of other agents (e.g., their cultural parents).  The underlying ``true'' environment, or world, is continually changing; a behavior that may be adaptive in one moment in time may no longer be advantageous if this world changes. The population has some average level of fitness which can be thought of as \textbf{the quality of a collective ``understanding'' of the world}.

In this world, individual learning is often considered costly and risky: the environment is stochastic, and the agent has some chance of failing to adapt to individual learning. Social learning may be cheaper and uncertainty-reducing, if many other agents have the same strategy then that may provide a signal of the quality of that strategy. However, social learning in a changing environment relies on some other agents having already successfully individually learned the adaptation and as a result, social learning is time-lagged compared to individual learning (providing a potentially outdated ``world model'' if the environment has changed). Intuitively, the availability of cheap social learning ought to be helpful for improving quality of collective world model by allowing efficient propagation of information; yet,  the cheapness creates an incentive for individual learners to become social learners. Accordingly, a relative shift in the proportion of individual learners in a network reduces the availability of timely new information. The fitness of the population at the resulting equilibrium is the same as if there were only individual learners meaning cheap social learning does not improve the collective world model: this is Rogers' paradox~\citep{rogers1988does}, which we re-instantiate through simulations, as discussed in Supplement~\ref{supp-base}.


\subsection{Background}

We next provide a concrete walkthrough of such a network. Consider the following thought experiment (setup adapted from \citet{deffner2022does} and notation adapted from \citet{enquist2007critical}). Suppose we have agents ($N=1000$) in a world that is slowly changing over time, with a fixed probability ($u=0.01$) that the optimal behavior for succeeding in this world changes at each time step (e.g., due to some weather event or other change to the affordances in the world). Agents in this environment who discover the optimal behavior (consistent with \citet{enquist2007critical} we call this the ``OK'' behavior) for a given timestep are ``adapted'' and have an increased probability of surviving ($s^{OK}:=P(\text{survival}|\text{OK})=0.93$) compared to non-adapted individuals ($s^{\neg OK}:=P(\text{survival}|\text{not OK})=0.85$). To discover the current optimal behavior, agents can attempt to learn about the current state of the world at a cost ($c_I=0.05$) and with some risk of failure (success probability $z_i:=P(\text{OK}|\text{individual learning}=0.66$, $p_i^{OK}:=(1-c_I)z_i=0.95*0.66=0.627$. At the end of each timestep, agents survive according to their survival probability, and the environment is replenished back to its original size with new agents. At the end of each timestep after this process, we measure the proportion of agents who are adapted ($q^{OK}$), and we track this quantity over many timesteps ($T=200000$) to find the long-run equilibrium fitness of the population: $E[q^{OK}]=p_i^{OK}s^{OK}$\footnote{In all simulations we explore in this paper, as discussed in the Supplement, the base $E[q^{OK}] = 0.58$.}. 

Now suppose that we introduce a second learning strategy into this environment: we allow some agents to learn socially by copying the behaviors they observe from another randomly selected agent. We assume social learning is much cheaper ($c_S=0$) and more reliable than individual learning, such that an agent is guaranteed to learn the optimal behavior if they observed another agent performing it and the environment has not changed that timestep, $p_s^{OK\rightarrow OK}:=P(\text{OK}|\text{observed OK})=P(\text{unchanged})P(\text{copied successfully})=(1-u)*1=0.99$. Thus, the probability of becoming adapted when social learning is equal to the proportion of adapted agents in the environment, $p_s^{OK}:=(1-c_S)q^{OK}p_s^{OK\rightarrow OK})$. Agents can either be individual or social learners (with proportions of $q_i, q_s=1-q_i$ of each type in the environment, respectively), but when new agents are added to the environment at the end of each timestep, they inherit their learning strategy from the surviving agents. Intuitively, we would expect that adding social learning, a cheap and reliable method of learning behaviors, would increase the average fitness; but surprisingly, the equilibrium reached by this network has the \textit{same} average fitness as when the agents only had access to individual learning. This is Rogers' Paradox \citep{rogers1988does}. We visualize this setup in Figure~\ref{fig:schematic}.

While initially unintuitive, Rogers' Paradox can be understood as arising from the \textit{tension} between individual learners having an incentive to become social learners when social learning has higher expected fitness ($E[q_s^{OK}]=E[p_s^{OK}]s^{OK}$) than individual learning ($E[q_i^{OK}]=p_i^{OK}s^{OK}$), and the expected fitness of social learners depending on the number of individual learners in the network.

\begin{equation} \label{eq1}
\begin{split}
E[q^{OK}] 
& = \frac{p_i^{OK}s^{OK}E[q_i]}{[1-(1-c_S)p_s^{OK\rightarrow OK}s^{OK}E[1-q_i]]}
\end{split}
\end{equation}



We can see in Equation~\ref{eq1} that the expected mean fitness across all agents in this case depends on the proportion of individual learners. The expected mean fitness of just social learners (Equation~\ref{eq2}) thus also depends on the proportion of individual learners.

\begin{equation} \label{eq2}
\begin{split}
E[q_s^{OK}] & = \frac{(1-c_S)p_s^{OK\rightarrow OK}s^{OK}p_i^{OK}s^{OK}E[q_i]}{[1-(1-c_S)p_s^{OK\rightarrow OK}s^{OK}E[1-q_i]]}
\end{split}
\end{equation}

As this network evolves over time, if the fitness of social learners is higher than individual learners, the proportion of individual learners will decrease until their propagation rates are equal resulting in the same mean fitness at equilibrium as when only individual learning was available -- even though both individual and social learners are present in the network. These results can be validated empirically by running simulations. We run simulations adapted from \citet{enquist2007critical} and \cite{deffner2022does} and visualize the results of these baseline simulations in Figure~\ref{fig:baseline} (left) and Supplement Figure~\ref{fig:orig-rp}. We provide implementation details in Supplement~\ref{supp-base}, as well as derivations.



\subsection{Introducing AI into the Network}

In this work, we extend the simulations to introduce an abstract ``AI model'' into the network. With the propagation of AI systems like GPT, we increasingly see people using these systems for knowledge retrieval, decision-making, and problem-solving - a clear form of social learning. However, these AI systems themselves are trained on virtually all the text ever produced by humans. Books, text, and the Internet can be thought of as cultural artifacts that reflect our collective world model; accordingly, AI systems can be thought of as simultaneously doing social learning from the entire population at once. The increased availability of these systems makes social learning from the AI system (and by virtue of the way the model is trained, the population therein) very cheap, and the fact that they could be viewed as reflecting a collective world model of the entire population maximizes the uncertainty-reducing effect of social learning; however, we hypothesize that these qualities alone are insufficient to resolve Roger's Paradox. 

\begin{figure}
    \centering
    \includegraphics[width=0.48\linewidth]{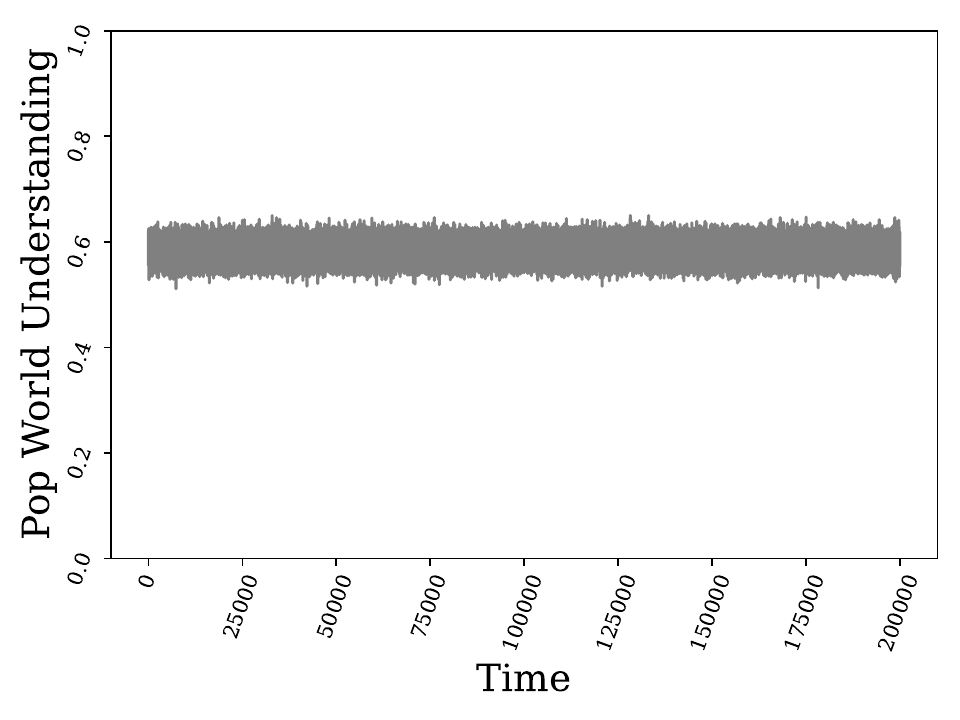}
    \includegraphics[width=0.48\linewidth]{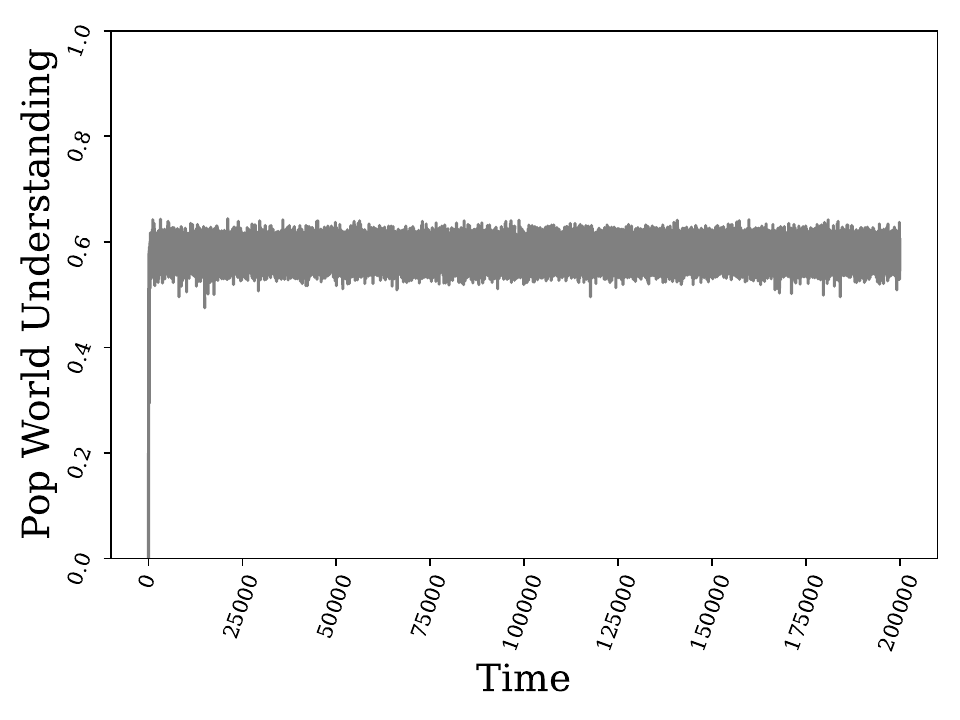}
    \caption{Comparing the collective world understanding over time, in a network where agents can only learn individually and do so at a cost (left) versus a network where agents can learn individually at the same cost or learn socially for free from an AI system that socially learns from all agents in the network (right). Each network attains the same baseline expected collective world understanding: recovering the classic Rogers' Paradox finding, even with an AI node.}
    \label{fig:baseline}
\end{figure}

We extend our simulations for studying Roger's Paradox to give agents in a slowly changing environment three choices: learn individually (has cost - $c_i=0.05$, may fail resulting in agent not adapting - $z_i=0.66$ - with overall success probability $p_i^{OK}=(1-c_i)z_i=0.627$), socially from a randomly selected other agent (no cost - $c_s=0$, succeeds only if the other agent is adapted - $p_s^{OK}=(1-c_s)q^{OK}p_s^{OK\rightarrow OK}$), or socially from an AI system (no cost - $c_{AI}=0$, succeeds with probability equal to adaptation level of the AI, $p_{AI}^{OK}:=q^{OK}$ - see below). At the end of each timestep, \textbf{the AI system learns socially from the entire population and matches the corresponding probability distribution of strategies (i.e., the AI's adaptation level is set to the mean adaptation status of the population)}. Agents are operationalized as being individual or social learners, and social learning agents have a propensity for learning from other agents versus from the AI system. Agents who successfully learn the correct strategy become adapted and have a slightly higher survival probability - so every timestep, the population-level propensity, for individual versus social learning and for social learning from agents vs from the AI, is updated. Running these extended simulations we again find that the average population fitness does not increase relative to the individual learning-only case (see Figure ~\ref{fig:baseline}). In fact, even if we decrease the relative cost of learning from the AI (by increasing $c_s$, the cost of socially learning from other agents) or increase the relative transmission success probability when learning from the AI (by decreasing $p_s^{OK \rightarrow OK}$) the mean population fitness at equilibrium (Equation ~\ref{eq3}) remains unchanged. 

\begin{equation} \label{eq3}
\begin{split}
E[q^{OK}] 
& = p_i^{OK}s^{OK}E[q_i] + E[p_s^{OK}]s^{OK}E[q_s] + 
E[p_{AI}^{OK}]s^{OK}E[q_{AI}]
\end{split}
\end{equation}

This suggests a novel form of Roger's Paradox for the AI age: the widespread availability of cheap AI systems trained on all human data in the world may not, on its own in the long-term, improve our collective world model. In the following section, we consider existing and novel directions in human-AI interaction and foundation model research through the lens of Roger's Paradox to evaluate their potential effect on our ability to leverage AI for improving our collective world model. We include additional details on the simulation environment in Supplement~\ref{supp-ai-base}.


\section{Exploring Strategies for AI Rogers' Paradox}
\label{sec3}

How can we design for positive learning outcomes when we introduce an AI agent into the network? There are multiple parties that can help make human-AI interaction ``go right'': the human, the developer of the AI model, and the developer of the infrastructure around the model (e.g., the interface builder, even ``society'' that may change affordances that guide use of an AI system). We consider several strategies that are motivated by the literature on human-AI interaction. For each strategy, we explore how one possible instantiation in our simulation framework may modify the collective ``world model'' acquisition and quality. 
We emphasize that our simulations act as a guide for imagining the impact that these strategies may have on \textit{population dynamics} (not just the \textit{individual} interacting with the AI system), as they have in evolutionary biology, anthropology, and other disciplines, extending to thinking about human-AI interaction. 




\subsection{Human- and Interaction-Centric Strategies}

We first consider strategies that can be undertaken by the individual human or addressed in the infrastructure around learning, e.g., by organizations, developers, or regulators.

\subsubsection{When Should You Learn from an AI System?}

Deciding who to learn from and when is not always an easy task with humans~\citep{schwartz2015paradox, kendal2018social}. Deciding when (and when not) to learn from an AI system is an ever more important question as these tools grow more powerful and accessible. In the context of human-AI interaction, there is a burgeoning literature for approaches that encourage the non-use of AI assistance in favor of human judgment~\citep{mozannar2020consistent,bhatt2024should,swaroop2024accuracy}. Within the network models around Rogers' Paradox, this equates to humans \textit{choosing} to engage with individual learning over social learning from the AI system. We as authors certainly advocate for critically appraising whether to engage with an AI system. We consider this idea in the context of our simulations by assuming that each agent can assess the relative expected utility of a given computation: accounting for their likely world knowledge versus that of the model, and the cost of individually boosting their knowledge versus engaging with the AI system. This appraisal could be done by each individual, but also may come with a cost (e.g., monitoring the quality of the AI system at test-time and assessing one's own abilities).

Recent work has highlighted that humans may not have a well-calibrated understanding of AI system capabilities ~\citep{vafa2024large} and may think a model's output is correct even when its not, perhaps arising from a lack of confidence in one's own ability~\citep{collins2024evaluating} or other ``illusions of understanding''~\citep{messeri2024artificial}. The \textit{infrastructure} around the human-AI interaction (e.g., the interface or other affordances that scaffold access to AI assistance~\citep{collins2024building}) can therefore be used to help an individual decide when they should engage with an AI system in social learning~\citep{collins2024modulating, li2024decoding}. One could imagine then that humans are told \textit{a priori} about the AI system's ability, quality, or cost and can use this information to decide where they should perform individual learning or social learning, which would correspond to engaging with the AI system. 

However, when we implement such a strategy -- by making the AI system unavailable when the expected adaptation value of learning from it is lower than the expected adaptation value of learning individually ($E[p_{AI}]^{OK}<E[p_i^{OK}]$) -- into our network model, we find that the population equilibria of world understanding do not change. This lack of a change to long-term equilibria from such a strategy is in line with other simulations extending the Rogers' Paradox formalism with strategies that turn out to improve individual but not collective fitness~\citep{boyd1995does}; we may imagine that introducing an AI system with different relative costs may change such network equilibria but do not find that is the case. While the research we discuss above fairly definitively identifies the \textit{individual} benefits of this type of strategy, here, we see that it does not translate into improvement in the \textit{collective} world understanding, but it does greatly reduce reliance on the AI system. Future work is well-poised to consider what collective impacts may arise from alternate ``who'' strategies, e.g., based on model transparency (see Section ~\ref{looking-ahead}).

\begin{figure}
    \centering
    \includegraphics[width=0.85\linewidth]{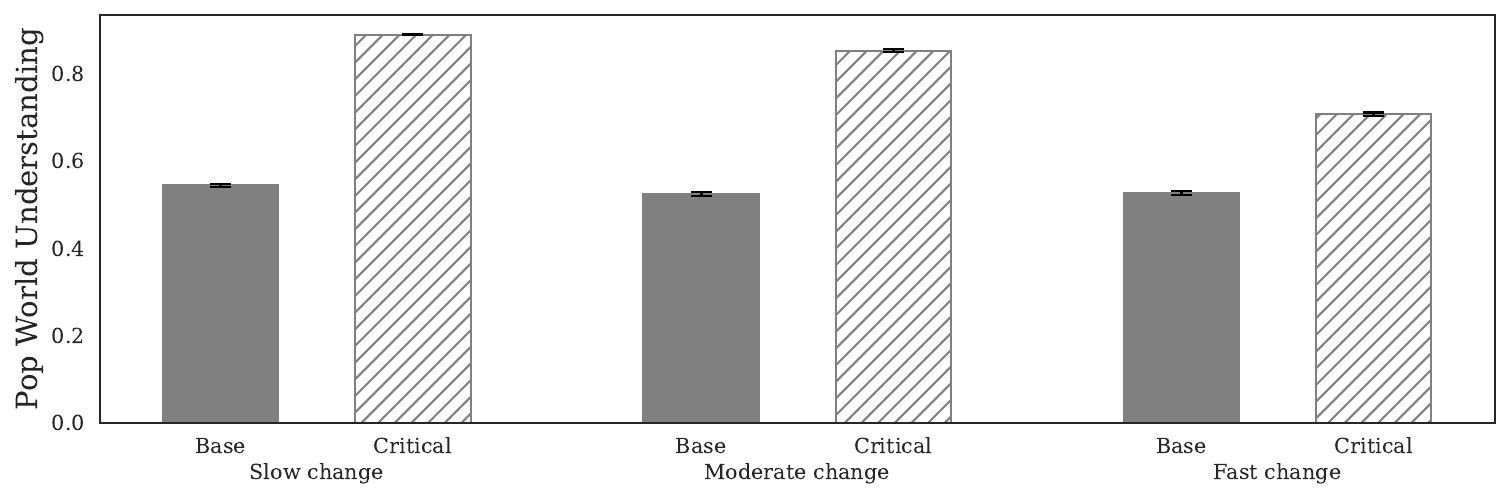}
    \caption{Impact of critical social learning (thatched) from the AI over the baseline learning strategy in the presence of AI (filled). Critical social learning leads to increased population world understanding, across varying rates of world change ($u=0.01, 0.1, 0.5$); however, critical social learning is a less powerful strategy if the world is changing very rapidly. }
    \label{fig:critical-social}
\end{figure}

\subsubsection{When Should You Override the Output of an AI System?}



After you have decided to engage with an AI system for learning, humans still have \textit{agency to decide} whether or not to uptake the system's output into their thinking. While the decision around \textit{whether} to engage with an AI system does not necessarily impact the equilibria of population world understanding in our current simulations, we next consider the impact of the decision to update one's knowledge of the world upon accessing the model.
Understanding when humans rely appropriately or inappropriately on the output of AI systems has been studied in the literature of behavioral economics~\citep{dietvorst2015algorithm,logg2019algorithm}, machine learning~\citep{mozannar2020consistent,guo2024decision}, and cognitive science~\citep{steyvers2022bayesian}. The preferences on when humans integrate the output of an AI system into their decision making will differ between individuals~\citep{bhatt2023learning,swaroop2024accuracy, steyvers2024three} and how humans are permitted to express an intervention to an AI system may depend on factors like the intervener's uncertainty~\citep{collins2023human}. In the same vein, policymakers are deciding when and how humans are given the ability to override a system~\citep{EUAIAct}. For instance, Article 14 of the EU Act explicitly spells out requirements for human oversight of AI systems: specifically, users must be granted the ability to understand the capabilities and limitations of a system, as it pertains to our transparency discussion above, and must be able to ``disregard, override, or reverse'' the outcome of an AI assistant. Similar requirements will emerge in the wake of Biden's Executive Order~\citep{EO14110} and Australia's 10 voluntary guardrails for AI safety that includes ``human control or intervention in an AI system to achieve meaningful human oversight''~\citep{au2024guardrails}.
Any human who uses an AI system for learning will be affronted with a decision on whether they integrate a system's output into their beliefs. We implement this in our simulations with a \textbf{critical social learning strategy}: a human can decide to access the AI system, and \textit{based on its output, can decide instead to switch to individually learning instead} ($p_{cs}^{OK}=1-(1-p_s^{OK})(1-p_i^{OK}$); ~\citep{enquist2007critical}). We find that a critical social learning strategy (see Figure \ref{fig:critical-social}) yields a \textit{higher} equilibrial state of collective world understanding. Simulation details can be found in Supplement ~\ref{supp-override}.


\subsection{Model-Centric Strategies}

We next explore some strategies that can be applied at the level of the model. a series of questions that warrant consideration to help guide the possible futures we laid out above in a positive direction. 

\subsubsection{How Often Should an AI System Update Its Understanding of the World?}

\begin{figure}
    \centering

    \includegraphics[width=0.7\linewidth]{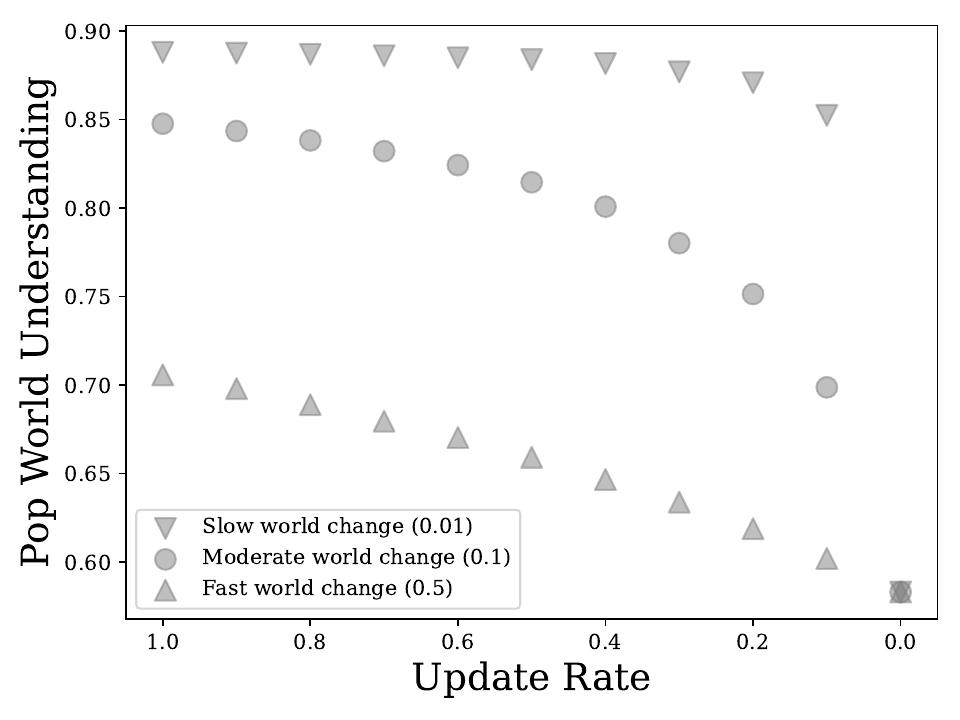}
    \caption{
    Impact of the update schedule of AI on the collective world understanding. Each dot represents the average population world understanding quality attained with the specific update schedule, in a world of a particular change rate. We consider three different rates of environment (world) change ($u$). Update rate signifies the probability that the AI will ``update'' (social learn) on any given timestep.}
    \label{fig:update-schedule}
\end{figure}

Our world is constantly changing. While humans can dynamically update their understanding of the world on-the-fly, the process by which an AI system can efficiently update its model remains in question~\citep{wong2023word}. In practice, however, there are very real costs to AI systems updating its understanding of the world~\citep{ibrahim2024simple, khetarpal2022towards, rillig2023risks}. This raises a question of strategy: in an ever-changing, interconnected world of learners, what is the impact of variable update schedules on collective world understanding? Thus far, we have assumed that the AI system snaps to the population mean on each iteration ($t$). We next consider the impact of the AI systems' ``update schedule'' (expected rate of AI engaging in social learning at each step; if AI currently has adaptation $x$, then $p_{AI}^{x\rightarrow q^{OK}}:=1-c_{\lambda_s}$ and $p_{AI}^{x\rightarrow x}:=c_{\lambda_s}$ where $c_{\lambda_s}$ is the cost of updating the AI socially to the mean of the agents in the network, so $p_{AI}^{OK,(t+1)}=1-(1-p_{AI}^{x\rightarrow q^{OK}}q^{OK})(1-p_{AI}^{OK,t}p_{AI}^{x\rightarrow x}(1-u)$).
We find in Figure \ref{fig:update-schedule} that there is a saturation point at which the frequency of updates has minimal impact on the equilibrium of the population's collective world model understanding (thereby, costs could be saved by a model builder from less frequent updates) that depends on the base rate of change in the environment. Yet, with too infrequent updates, we see deleterious impacts on the equilibria of collective world understanding. These impacts are exacerbated in a base world that is changing more rapidly, rendering learned knowledge and behaviors quickly obsolete, as we see in Figure \ref{fig:update-schedule}. Simulation details can be found in Supplement~\ref{supp-update}.


But, when the world is changing continually, not all changes in the environment are worth necessarily triggering an update, if there is some cost to each update. There are likely many aspects of the world that do not warrant change when they occur. If it snows in Boston in January, we may not need to go through the costs of updating an AI system (we assume the model has some ``understanding'' that snow is a regular occurrence in Winter); however, if it snows in Boston in June, the salience of the observation may demand an update as it indicates that something fundamentally different about the world has been observed. Likewise, we may imagine that not all interactions with users warrant an update, or inclusion in the update set (e.g., the preferences used for any kind of human feedback based fine-tuning~\citep{ouyang2022training, christiano2017deep}). Future work can explore the network effects of salience-based updates or alternate more realistic scheduling protocols, as well as accounting for the possible impact on the efficacy of the human-AI interaction from an update~\citep{bansal2019updates}.

\subsubsection{How Should an AI System Update Its Understanding of the World: Socially or Individually?} 


While many popular AI systems today (e.g., several large language models) may be viewed as having socially learned from us, the quality of such an AI system's ``understanding'' of the world is then based on the knowledge of the individual humans in the system: the advantage to the population comes from the AI system's ability to consolidate and provide access to this knowledge at a cheap cost to other humans, not ``new'' knowledge per say. To go beyond what humans know to bring our collective ``world model'' closer to the actual world, we may also consider an AI system which updates based on its individual interactions with the world. Many researchers in AI have, and are, pursuing alternate training schemes, wherein AI systems learn from non-human generated data (e.g., distilling information from from other AI models~\citep{hinton2015distilling}, engaging with explicit simulators of the world~\citep{toolsGame2020, groundedLanguageLearning2017}, or relying on self-play~\citep{silver2017mastering}). Recent approaches have begun examining test-time compute scaling as an additional way to improve AI performance (albeit at a fairly steep cost) by letting the AI `think' for extended periods of time before responding \citep{snell2024scaling}. Increasing discussion around ``agentic'' AI also opens up the possibilities that AI systems individually learn about the world by taking action in that world, for instance, iteratively checking code against a compiler, searching the web, or taking physical action. 

\begin{figure}
    \centering

    \includegraphics[width=0.49\linewidth]{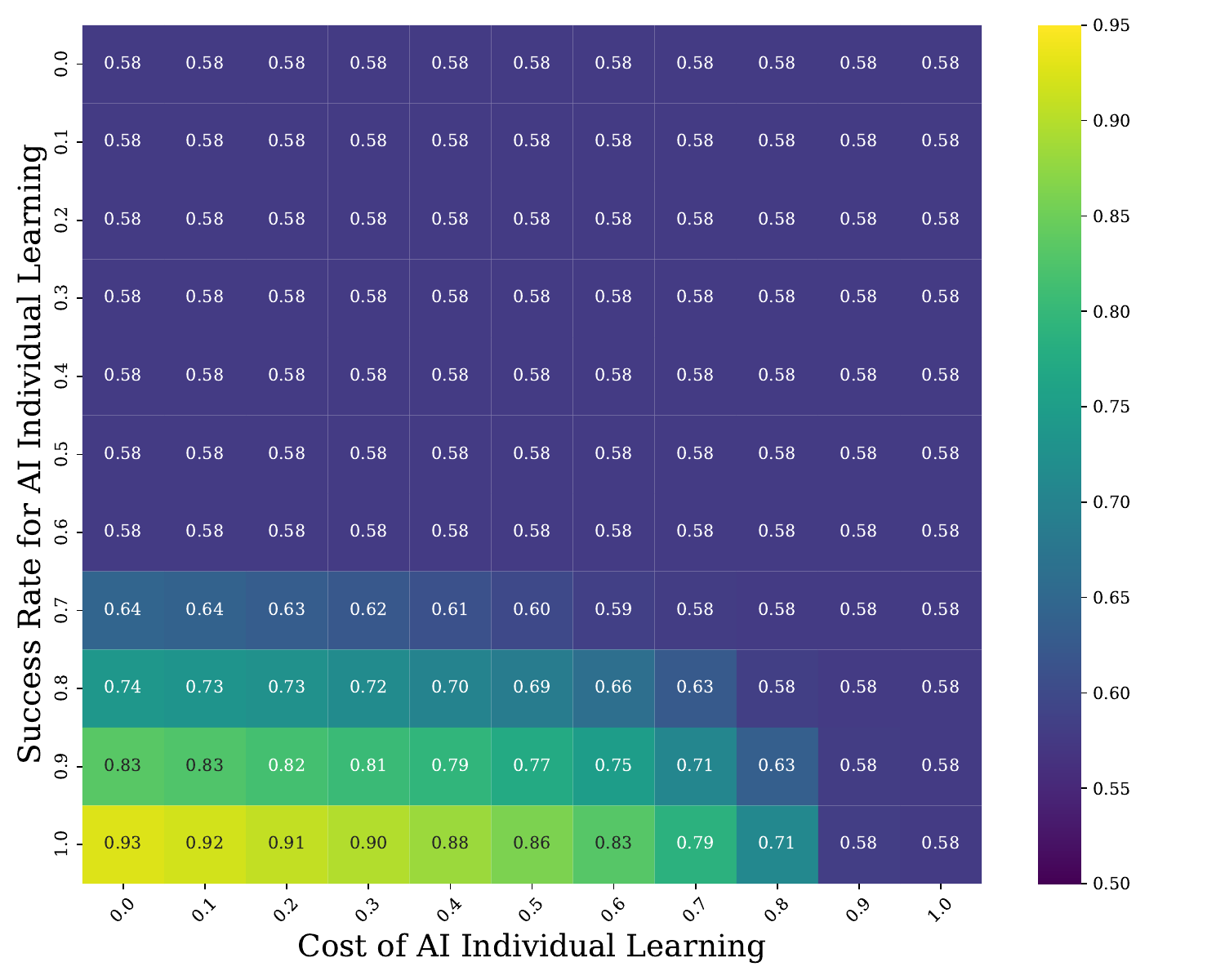}\includegraphics[width=0.49\linewidth]{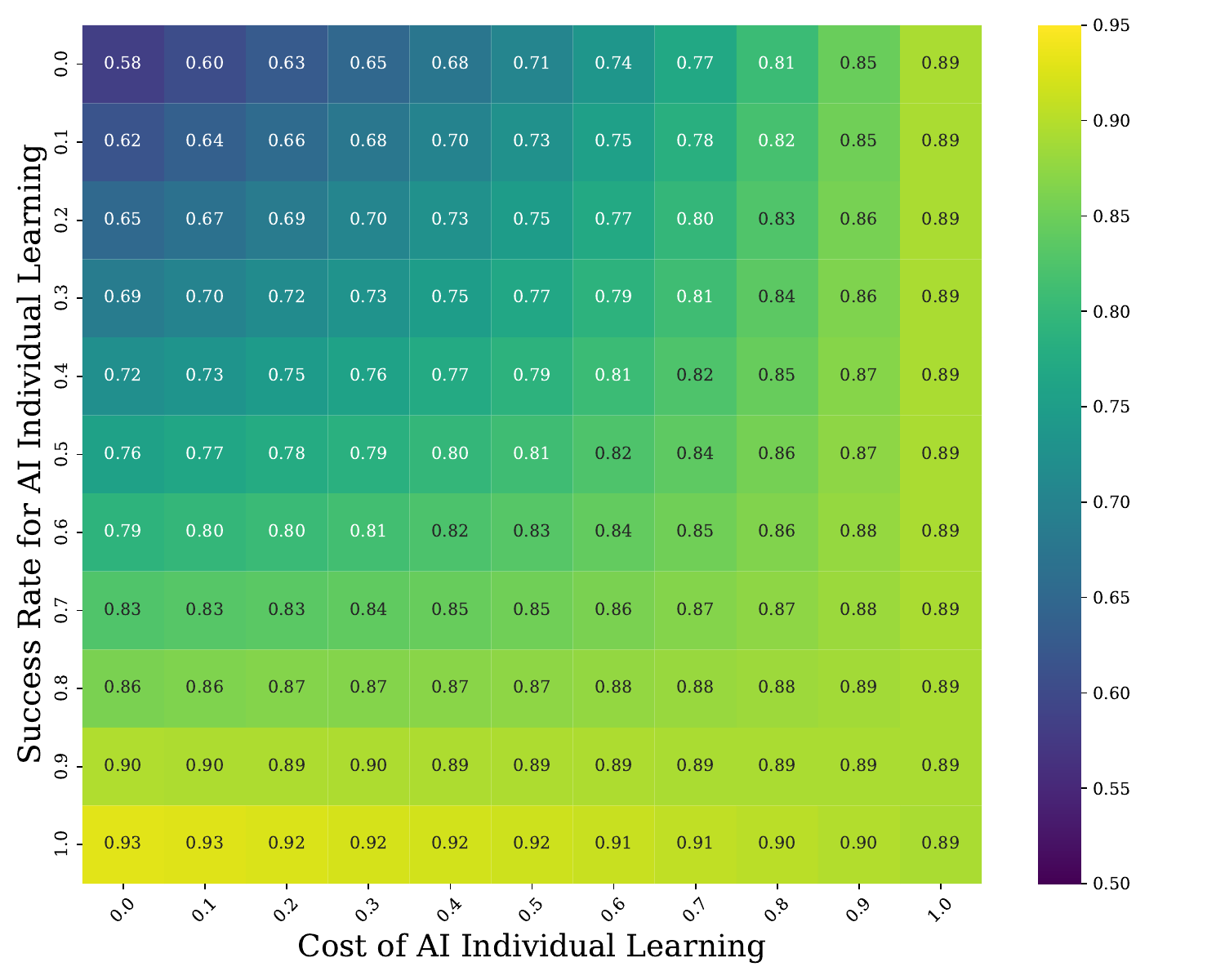}
    
    \caption{Allowing the AI system to either individually learn or socially learn on each turn. Impact on collective world understanding (color of cells), depending on the cost to the AI individually learning (x axis) and the expected quality of the AI individually learning about the world (y axis). Left: baseline network wherein agents either individually learn or socially learn from the AI; Right: network where agents critically social learning from the AI. 
    }
    \label{fig:ai-individ-update}
\end{figure}

Whether an AI system has learned socially (from us) or individually (from self-play or other actions), we may learn from such an AI system through the same mechanisms. We explore this idea by considering an AI agent in our population network which can learn individually, with some cost relative to social learning ($p_{AI}^{x \rightarrow OK}:=(1-c_{\lambda_i})z_{AI}$ where $c_{\lambda_i}$ is the cost of updating the AI individually, and $z_{AI}$ is the success rate for the AI's individual learning). But in practice, just because an agent can individually learn does not mean that such learning will be successful. We take first steps to explore the impact of varied individual learning success rates and varied individual learning costs on population dynamics in Figure~\ref{fig:ai-individ-update}. We see that when the AI system has a low cost to individually learning and a high success rate when pursuing individual learning -- the population's net collective understanding of the world can improve substantially, regardless of whether the population (or infrastructure around users' interactions with the AI) involves critical evaluation of use. However, we see that when the AI is often \textit{unsuccessful} with individual learning (i.e., it comes to a bad conclusion about the world) and particularly when costs to individual learning are low such that the model is frequently individual learning, then population world understanding is hampered, unless the human agents have not already shifted to critically appraising the model's output. These data further drive home the intuition that if the AI's individual learning is unsuccessful, it is all the more crucial that an individual human, or infrastructure around their access, e.g., including transparency, critically assess whether it is worth uptaking the output of an AI system. Notably, we see that the \textit{composition} of strategies is especially crucial when the AI system either has a higher cost to individual learning, or lower success rate: it remains important that the population engages in some form of critical appraisal on whether or not to override the output of the AI system. Exploring the relative benefits of compositions of human- and model-centric strategies demands further study, especially as it may inform whether it is worth ``paying'' for more particular kinds of AI updates or modes of learning. These questions grow ever more pertinent when considering potential relative risks of ``agentic'' AI systems~\citep{chan2023harms}. Additional simulation details can be found in Supplement~\ref{supp-ai-individ-learn}.

\section{When Interactions Change Learning Efficacy}
\label{negative-feedback}

Now, imagine if you could at the snap of a finger engage instantly with an expert to learn about the world? What if you never needed to pour over a textbook for hours on your own to learn a new concept -- to get stuck and need to start again? While to (hopefully) many, this would not be a particularly fun or fulfilling future: 
if you can always socially learn for cheap and therefore ``avoid'' individual learning, your \textit{future ability} to re-engage with individual learning may be substantially weakened. Thus far, we have focused on the impact of various learning strategies in a modified version of the original Rogers Paradox setting wherein a human attempting to improve their understanding of the world can choose to learn individually or socially from another agent (human or AI). However, this choice has no impact on the human other than improving (or failing to improve) that person's understanding of the world. Whichever way you choose to learn, you will always have the same expected learning success. Yet, the ways that you choose to learn \textit{can} impact how successful your future learning may be. This is especially a concern with AI tools or possible ``cognitive extenders''~\citep{hernandez2019ai}. Several works have raised concerns about the impact of AI tools on our cognition and relative self-appraisal: these include algorithm appreciation~\citep{logg2019algorithm}, loafing~\citep{inuwa2023algorithmic,saluja2024loafing}, algorithm aversion~\citep{dietvorst2015algorithm,dietvorst2018overcoming}, algorithmic vigilance~\citep{zerilli2022transparency}, de-skilling~\citep{rafner2022deskilling,glickman2024human}. We therefore next introduce a slightly richer network of agent interactions into our simulations: \textit{negative feedback upon interaction}.

\begin{figure}
    \centering
    \begin{minipage}{0.32\linewidth}
        \includegraphics[width=\linewidth]{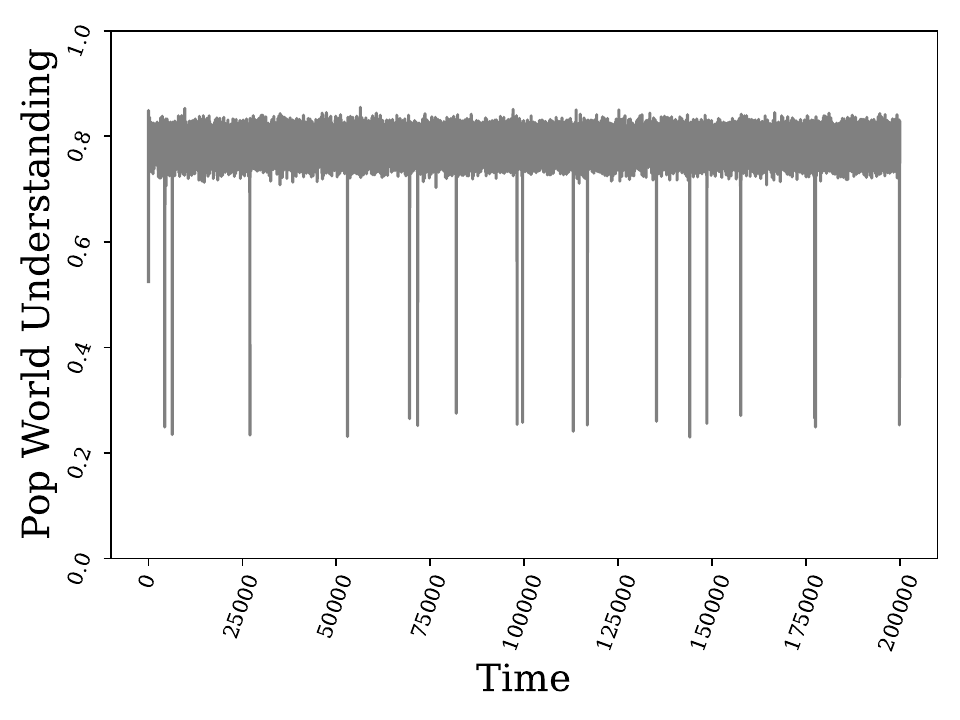}
    \end{minipage}
    \hfill
    \begin{minipage}{0.32\linewidth}
        \includegraphics[width=\linewidth]{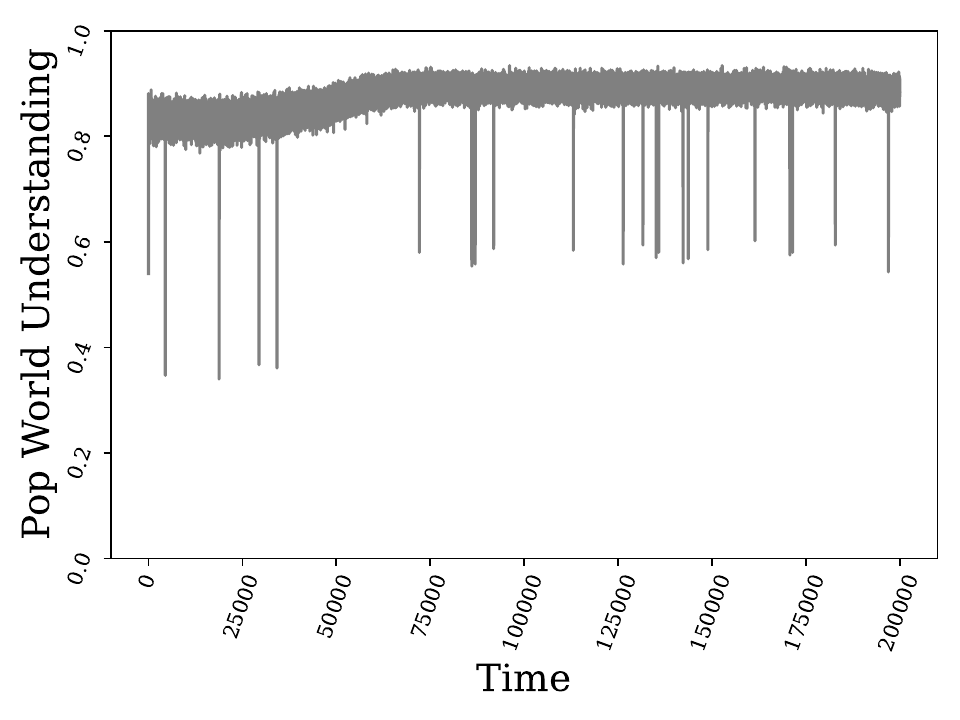}
    \end{minipage}
    \hfill
    \begin{minipage}{0.32\linewidth}
        \includegraphics[width=\linewidth]{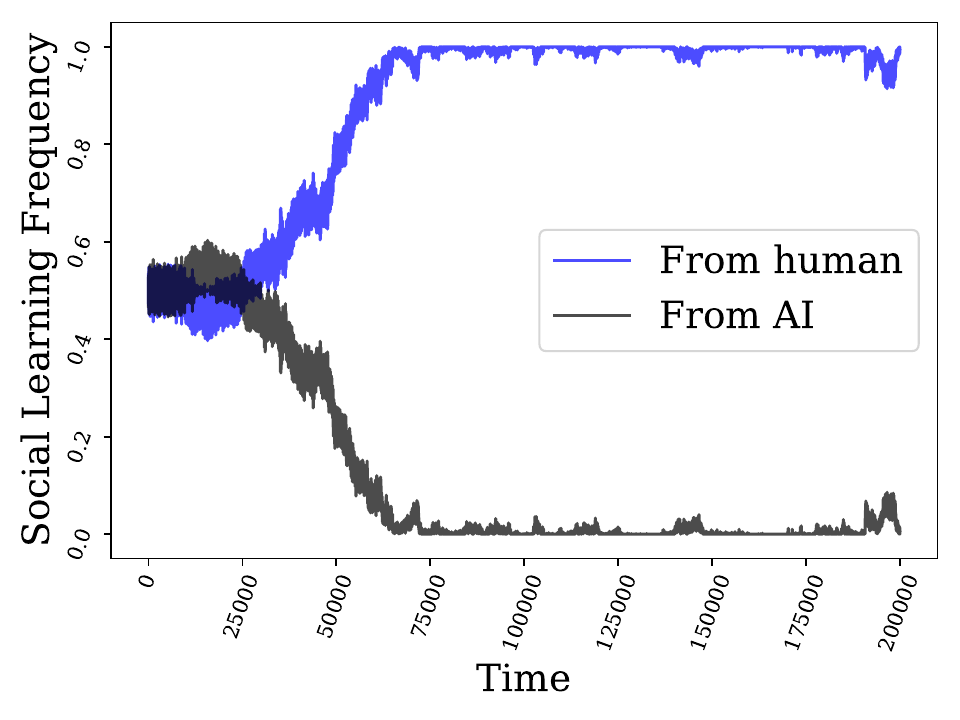}
    \end{minipage}
    \caption{\textbf{Left}: Critical social learning with access only to AI and there is negative feedback (learning from AI makes you worse at individual learning). 
    \textbf{Center}: Critical social learning with access to both AI and humans (with negative feedback from AI). \textbf{Right}: Learners start in the low equilibrium but phase out the AI after a period of time to reach the high equilibrium.}
    \label{fig:negative-feedback}
\end{figure}

The simulation framework spawned from Rogers' Paradox offers a fruitful environment to explore the impact of this kind of negative feedback-induced deskilling. We take a step toward exploring this idea by implementing a negative feedback loop: when a human chooses to learn socially from the AI system, the success rate of their own individual learning decreases ($\kappa_{j}^0=1, \kappa_{j}^{t+1}=0.9\kappa_j^{t}$ for agent $j$ who learns from the AI, $p_{ij}^{OK}=z_i(1-c_i)\kappa_j$) . To our knowledge, this is a new innovation on top of the base Rogers' Paradox, whereby engaging with another entity in the network change the \textit{learning efficacy} of a given agent. We see in Figure~\ref{fig:negative-feedback} (left) that when there is negative feedback from learning from the AI, the equilibrium collective world understanding dips below that of baseline critical social learning. This makes sense: access to another agent who makes you worse at your own ability to learn (and who too is influenced by your understanding of the world) will run themselves downward to a globally poorer equilibrium. However, in Figure~\ref{fig:negative-feedback} (middle), we see that allowing agents two options for social learning: a more expensive human versus a cheaper AI (which degrades your own ability to learn) can maintain high collective world understanding. As depicted in Figure ~\ref{fig:negative-feedback}, the population \textit{adapts} to \textit{which source} to learn from socially. Over time, the agents determine that they are better off learning from the agent that does not weaken their own ability to individually learn when they need to. We could imagine more intricate and realistic extensions upon this framework, wherein an agent may shift from negative to positive reinforcement: if so, humans may benefit from a signal (e.g., a nudge~\citep{thaler2009nudge}) that it is worth learning from the AI to sidestep algorithmic aversion~\citep{dietvorst2015algorithm}. Details on our negative feedback simulation can be found in Supplement~\ref{supp-feedback}.



\section{Looking Ahead}
\label{looking-ahead}
We close by noting several open directions that excite us about studying human-AI interaction in the context of collaborative learning in these kind of network models, which we believe are ripe for richer representations of our uncertain, dynamic world. To support further exploration of and around AI Rogers' Paradox, we make all simulation code open-source at the following repository: \href{https://github.com/collinskatie/ai-rogers-paradox}{https://github.com/collinskatie/ai-rogers-paradox}.


\subsection{Handling Different ``Classes'' of AI Systems}

Our analyses thus far have focused on a single aggregate, abstract ``kind'' of AI system. However, at the time of writing, there is a burgeoning offshoot of more traditional large language models that undertake more compute at inference-time returning an output to the human (e.g., o1 ~\citep{jaech2024openai}). These models may come with higher costs for the human to learn from (e.g., longer time for a response; higher energy costs to providing the output), yet also produce a ``higher quality'' response. Just as humans then often make decisions about which humans to learn from~\citep{schwartz2015paradox, kendal2018social}, we may increasingly have a second order question of not only when to engage with an AI system for learning about the world --- but \textit{which} AI system to engage with. This puts an even stronger emphasis on the human being able to appraise the relative cost of their individual learning with the costs of engaging with an AI system, and being able to understand the expected utility of the different model classes.



\subsection{Transparency and Information Communication}
Humans frequently employ explanations to justify our recommendations and choices when learning from each other. Our analyses here do not account for (1) what potential information humans can access about AI systems, and (2) how access to that information affects their decision to engage in social learning.
More broadly, AI systems may communicate their understanding of the world to humans in an attempt to modulate when and how humans integrate an AI system into their learning process. There is a large literature on the transparency of AI systems, which could potentially be repurposed for world model communication~\citep{gunning2019darpa,ehsan2021expanding}. Much of this work refers to obtaining and designing explanations of the behavior of an AI system in humans~\citep{doshi2017towards}. The goal of such transparency is often to modulate how and when humans elect to learn from an AI system~\citep{zerilli2022transparency}. This transparency information could span from the communication of uncertainty information~\citep{bhatt2021uncertainty} to natural language explanations of behavior~\citep{zaidan2007using} to representations of the systems' world model via explicit probabilistic programs~\citep{vidal2022explanations, wong2023word, collins2024building, dalrymple2024towards}, or other indications of each agents' representation of the world as it relates to how that agent may communicate with another agent~\citep{sucholutsky2023getting,sucholutsky2024representational}. Thus far, our network simulations have focused on a single determiner of success: whether you can ``play'' the right strategy or not, based on your understanding of the world. Future work could separate out these components of deciding what to do and what is known about the world, as they relate to what you know about another agent. Transparency into an AI system's understanding of the world may also be used to deter from social learning~\citep{dietvorst2015algorithm, zerilli2022transparency}. In the settings we have considered here, the ``use'' of an AI system is resigned in favor of human judgment and individual learning based entirely on the relative success probabilities of ``playing'' a certain learned strategy. In our current instantiation of a form of selective use, we have assumed humans are only provided with the option to see and disregard the AI system output; transparency information instead permits the communication of \textit{why} an output was shown or not shown (e.g., the AI system's understanding of the world is knowingly flawed, e.g., due to infrequent updates or the uncertainty is too high). Although our simulations find that such interventions do not affect population equilibrium, future work can account for alternate veils of transparency and known confounding effects for how humans update their decisions to learn upon receiving information from an AI system during learning~\citep{buccinca2021trust,bhatt2023learning}. We may also consider varying degrees of information richness passed between agents about the world, which could include, for example, soft labels that capture a human's uncertainty on the task at hand~\citep{collins2022eliciting, sucholutsky2023informativeness}.

\subsection{Collaborative Learning with Human-AI Thought Partners}

The bulk of analyses related to Rogers' Paradox focus on agents as ``lone explorers'' engaging with the world, or agents learning hierarchically from another agent (akin to a child learning from a parent or student learning from a teacher). However, humans also learn about the world by \textit{partnering} with each other~\citep{Tomasello_Kruger_Ratner_1993} or other forms of ``peer learning''~\citep{lew2023peer}. Much good science has been done by two or more scientists thinking together~\citep{yanai2024takes} and even interaction between a teacher and student is not unidirectional~\citep{chen2024hierarchical}. Partnering often involves some kind of bidirectional \textit{intentionality}: reasoning about the others' goals and what they know and do not know~\citep{tomasello2005understanding}. Humans and AI systems though do not necessarily have the same structure to any kind of ``world model'' ~\citep{vafa2024world, mitchell2023ai}, ability to communicate understanding in mutually compatible ways ~\citep{kleinberg2024inversion, collins2024building,liu2024large}, or costs around modes of engagement. What possibilities (and risks) may arise for different flavors of co-learning agents? One could imagine extending the negative feedback modifications we made in Section~\ref{negative-feedback} to explore positive feedback, e.g., where co-learning with an AI system may boost an agent's ability to individually learn and vice versa. 



\subsection{Lowering Barriers to Human-Human Social Learning}


Our simulations also spotlight the question around who you choose to partner with and when: if your goal is to develop a faithful understanding of the world, when do you engage in partnering over vertical social learning versus individual learning? Presently, good human partnering is partly limited by access: in science, it can be hard, though a gift, to find good collaborators. And the human partners that we find may generally be ``like'' us in many ways, likely a byproduct of how we found the partner in the first place (e.g., same lab or university). The introduction of AI systems into networks of human learners may not only impact collective world model building by increasing the ease of access of good AI-based thought partners to human learners to ameliorate challenges of finding good human thought partners --- AI systems in these networks can also make it \textit{easier} to engage with a wide range of human thought partners. For instance, humans have, to an extent, been limited by how many other humans we can engage with at the same time in learning. AI systems, however, raise the prospect that humans can actually learn more efficiently from many humans. For instance, an AI system could summarize many different humans' responses, as in ~\citep{tessler2024ai}, allowing humans to consider the thoughts of many other humans more quickly. Future work can explore incorporating other actions of AI systems (e.g., as summarizers) in these agent-based population models. 



\subsection{Environmental Change and AI}

The above possibility (introducing AI systems to change the relative costs of human interaction) underlies another important direction of study building off the Rogers' Paradox-inspired network models we consider here: what happens when an AI system changes the environment by its ``presence''? Thus far, we have considered how the introduction of AI systems into networks of human and AI is impacted by changes in the environment -- when we assume unidirectional impact from environment change on the world model of an agent. However, it is possible -- and one may argue, even the current state of society -- where changes from the AI system \textit{change the rate of change in the environment}. Future work could explore the network effects of such dynamics. 

\subsection{Revisiting our Definition of ``Fitness'' and Notion of a Collective World Model}

We close by noting an important assumption that has been baked into our discussion thus far: that there is a ``single collective world model'' that can be learned and that society's overall ``fitness'' is based on the quality of this learned world model. This is sensible in the context of our coarse environment model: there is only ``one'' world with an oracle strategy for agents to deploy in this world. However, human societies are immensely diverse, and it is highly simplistic to assume that AI systems and any one human easily learn a good model of the world that captures this diversity. More likely, a collective world model will disproportionately represent some experiences and cultures over others, and an ``aggregate'' fitness metric washes out much of this potentially heterogeneous dispersion of understanding~\citep{kleinberg2021algorithmic, kirk2024prism}. At present, our simulations have assumed that AI systems learn from the mean of the population of learners -- but in practice, AI systems (often) learn from whatever data is most available, which regularly is enriched with particular biases~\citep{kotek2023gender,salinas2023unequal}. Data collection fails to represent the global majority, who may increasingly begin to interact with and learn from AI systems in the coming half century~\citep{zha2023data}.  We urge caution in interpreting these kind of simple network models, using them to fan intuition nad further thinking, rather than a rote guide. We also offer a hopeful note that one positive use of AI systems in networks of learners, as we discuss above in the context of lowering barriers to learning from other humans, is that AI systems may enable us to scale and diversify our understanding from a broader space of humans with whom we may not otherwise have engaged.


\section{Conclusion}

Rogers' Paradox spawned a rich line of work exploring the relative benefits of individual versus social learning among networks of humans in a dynamic, uncertain world. The introduction of powerful AI systems, which learn from us and which we now may increasingly learn from, opens up new questions about the ripple effects on our collective understanding. Our preliminary network simulations shed light on the importance of critically appraising whether learning from an AI system is sensible -- where that appraisal can be made by a human, the AI model builder, or introduced through other scaffolding around the moment of the human-AI interaction (e.g., in the design of interfaces, or regulation). We offer a new extension of these simulations that also models the potential of an AI system impacting the efficacy by which we can in turn individually learn about the world. However, the simulations we introduce and extend here are necessarily simplistic and only a first step. Future work is well-served to explore the impacts of other modifications to such network models (e.g., introducing multiple AI systems with different costs, or a wider range of possible behaviors in the world). AI systems are increasingly transforming our understanding of the world and each other. The coming years demand work work across multiple levels of abstraction and fields to understand the requirements for, and effects of, deploying evermore capable AI systems into our cultural fabric. 

\section*{Acknowledgments}

We thank Mark Ho, Danny Collins, Lance Ying, Adrian Weller, Josh Tenenbaum, Tom Griffiths, Jose Hernandez-Orallao, Nori Jacoby, and Kerem Oktar for valuable discussions that informed this work. KMC acknowledges support from the Cambridge Trust and King's College Cambridge. This work is supported (in part) by ELSA - European Lighthouse on Secure and Safe AI funded by the European Union under grant agreement No. 101070617. 
Views and opinions expressed are however those of the author(s) only and do not necessarily reflect those of any of these funding agencies, European Union or European Commission. 

\bibliography{main}
\bibliographystyle{abbrvnat}

\newpage 

\appendix

\section*{Additional Details on Base Simulations}

We include additional details on our simulations. All code will be open-sourced upon publication at the following repository: \href{https://github.com/collinskatie/ai-rogers-paradox}{https://github.com/collinskatie/ai-rogers-paradox}.

\subsection*{Base Network Model}
\label{supp-base}

We first set up social learning simulations in the style of the original Rogers' Paradox designs, as discussed in Section~\ref{sec2}. We simulate 1000 agents in a slowly-changing, stochastic environment (P of 0.01 that the environment changes at each timestep). Agents who are adapted to the environment at each timestep have a slightly increased chance of continuing to the next timestep (P of not being eliminated 0.93 instead of 0.85), but when the environment changes, all agents become maladapted until they learn the new strategy. We consider two types of agents: individual learners and social learners. At each timestep, depending on their type agents can either attempt to learn individually to try to adapt to the environment, or to learn socially and copy another agent's strategy. Learning individually from the environment has a cost and a risk of failure (P of success = 0.66) but if successful,  provides up-to-date information on the environment making the agent adapted. Learning socially from another (randomly selected) agent is free, but the learner only becomes adapted if the teacher was adapted, relying on potentially stale information from a previous round. At the end of each timestep, the population is replenished back up to 1000 agents with new agents descended from the remaining ones with a small mutation rate causing them to flip their learning strategy (P of mutation is 0.005).

\begin{figure}[h!]
    \centering
    \includegraphics[width=0.5\linewidth]{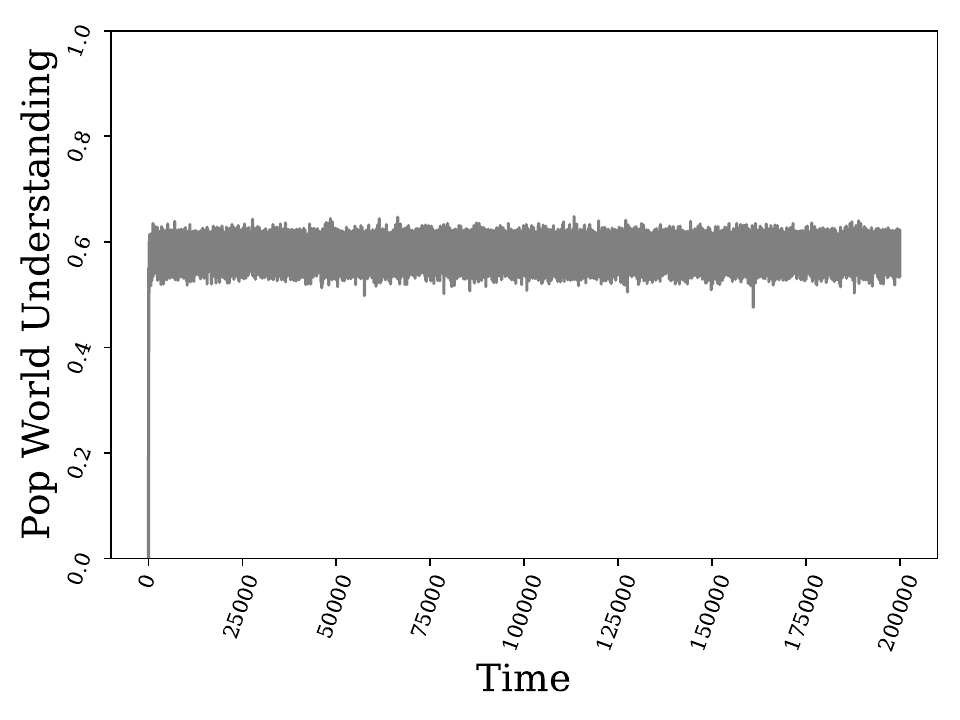}
    \caption{Quality of population world understanding in the original baseline Rogers' Paradox setting: only individual or social learning from other humans allowed. This is identical, in equilibrium, to a network of only individual learners, or a network of humans and an AI social learner.}
    \label{fig:orig-rp}
\end{figure}

We run one simulation where social learning is unavailable and there are only individual learners, and a second simulation as described above where both social and individual learners are present (see Figure~\ref{fig:orig-rp}. We run each simulation for 200,000 timesteps and measure the proportion of adapted agents at each time step. We average the proportion over the final 50,000 steps to get the population fitness (collective world ``understanding''). 

We also include a derivation of expected fitness, as introduced in Section \ref{sec2}, including the expected mean for all agents: 

\begin{equation} \label{eq1}
\begin{split}
E[q^{OK}] 
& = p_i^{OK}s^{OK}E[q_i] + E[p_s^{OK}]s^{OK}E[q_s] \\
& = p_i^{OK}s^{OK}E[q_i] + (1-c_S)p_s^{OK\rightarrow OK}s^{OK}E[q^{OK}]E[1-q_i]\\
& = \frac{p_i^{OK}s^{OK}E[q_i]}{[1-(1-c_S)p_s^{OK\rightarrow OK}s^{OK}E[1-q_i]]}
\end{split}
\end{equation}

And the expectation for just social learners:

\begin{equation} \label{eq2}
\begin{split}
E[q_s^{OK}] & = E[p_s^{OK}]s^{OK}\\
& = (1-c_S)p_s^{OK\rightarrow OK}s^{OK}E[q^{OK}]\\
& = \frac{(1-c_S)p_s^{OK\rightarrow OK}s^{OK}p_i^{OK}s^{OK}E[q_i]}{[1-(1-c_S)p_s^{OK\rightarrow OK}s^{OK}E[1-q_i]]}
\end{split}
\end{equation}
    

\subsection*{Introducing AI to the Network}
\label{supp-ai-base}

\begin{figure}
    \centering
    \includegraphics[width=0.5\linewidth]{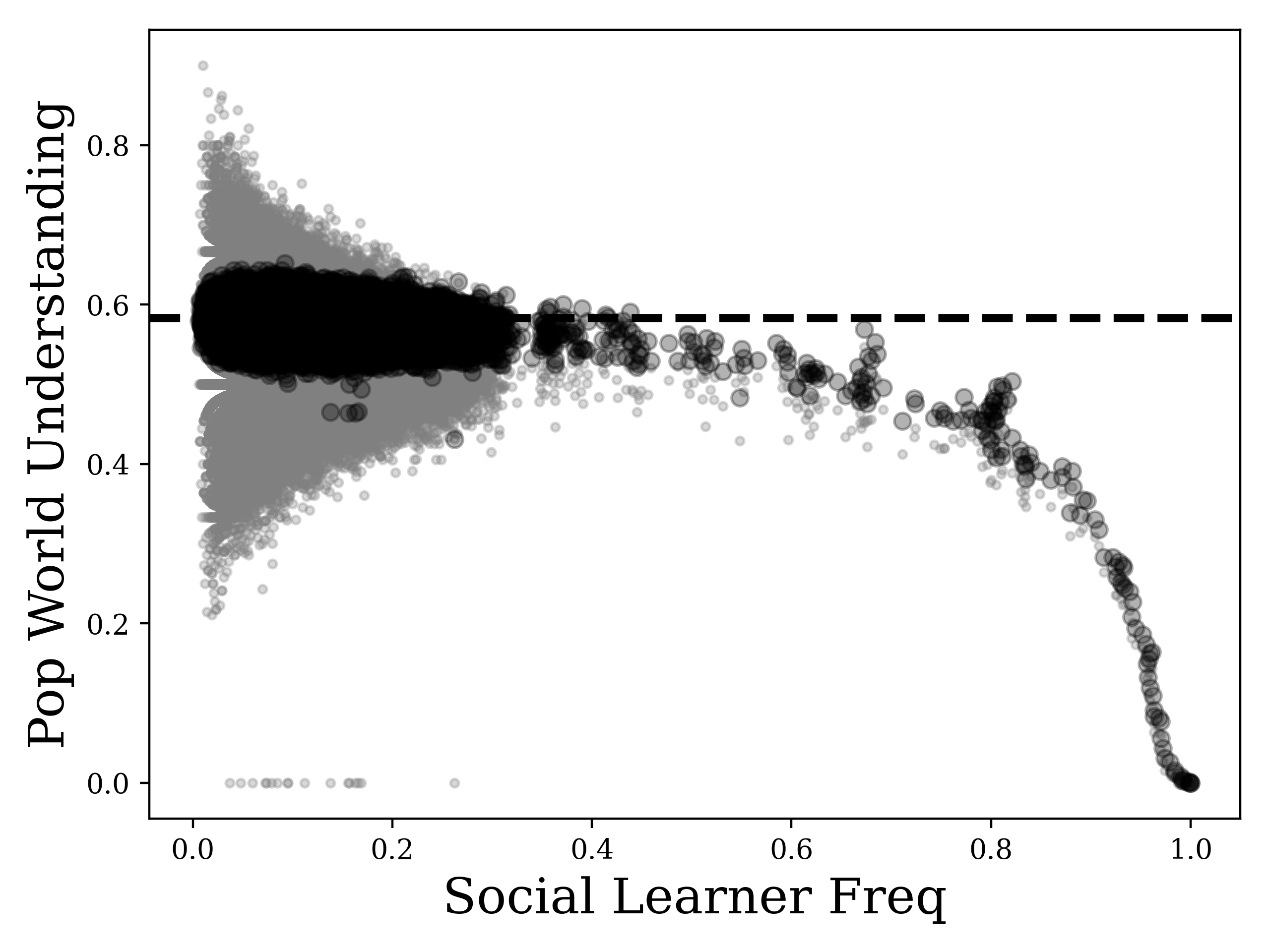}
    \caption{Social learning frequency versus population world understanding for the baseline AI Rogers' Paradox case from Section \ref{sec2}. The dashed line indicates the equilibrium of collective world understanding from a network of purely individual learners. Black indicates mean understanding of all learners; gray indicates mean of just social learners. When expected value of social learning exceeds that of individual learning, individual learners become social learners and vice-versa, resulting in the same equilibrium as in the individual learning case.}
    \label{fig:baseline-ai-social-learning-dynamics}
\end{figure}

We extend the base simulations to modify how the AI system learns at each step. The AI system either learns by snapping to the mean of the population, or ``pays'' a cost of individually learning. We vary the cost of individual learning and expected accuracy of the AI system individually learning. We assess the relative gains in the base Rogers' Paradox scenario (when humans either individually or socially learn based on their reliance propensity) versus a population wherein people engage in critical learning. We illustrate additional network dynamics in Figure~\ref{fig:baseline-ai-social-learning-dynamics}. We use the same parameters controlling world dynamism, human preferences, and human learning quality as in the previous sections. 

\section*{Additional Details on Strategy Simulations}

\subsection*{When Should You Learn from an AI System?}
\label{supp-you-learn}

\begin{figure}
    \centering
    \includegraphics[width=0.5\linewidth]{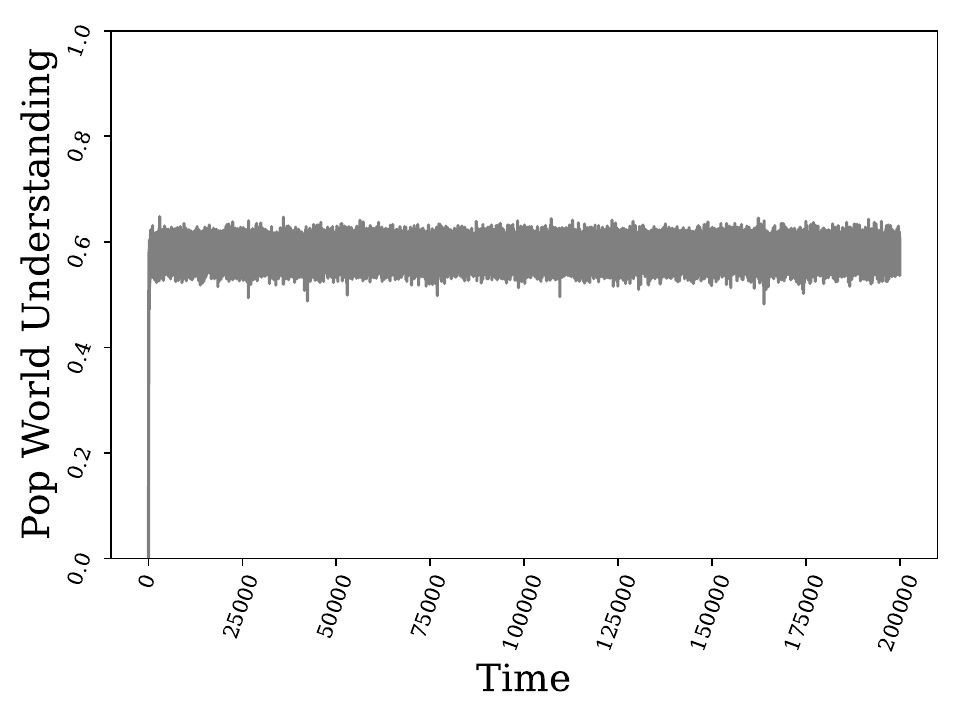}
    \caption{Impact of permitting an agent to decide whether to learn individually or socially from the AI; there is no net impact on collective world understanding.}
    \label{fig:alg-resignation}
\end{figure}

We assume that the individual learning success probability is known, and performance of AI system (i.e., conditional probability of success when learning from AI assuming the environment has not changed) can be evaluated. We extend the simulation to consider the case where social learning from the AI is only available when the probability of becoming adapted after learning from the AI exceeds the probability of becoming adapted after learning individually. This can be considered a form of conditional social learning where the agent always attempts to social learn, unless it is known that social learning is unlikely to succeed relative to individual learning. Unlike typical conditional social learning strategies, here the agent decides whether to social learn or individual learn \textit{before} seeing the output of the teacher (in this case the AI). This particular strategy has no net impact on the equilibria of collective world understanding, as depicted in Figure~\ref{fig:alg-resignation}.

\subsection*{When Should You Override from an AI System?}
\label{supp-override}

We next imagine that the agent can tell whether they became adapted after attempting to learn: that is, the agent can \textit{evaluate} how successful a learning interaction was, in line with~\citep{enquist2007critical}. We extend the simulation to consider the case where agents can override the outputs of the AI system by learning individually if social learning first fails. All other parameters are the same. 


\subsection*{How Often Should an AI System Update Its Model of the World?}
\label{supp-update}

Rather than having the AI system socially learn on every iteration, we consider a variable update schedule where the model socially learns with probability $u$ on each iteration. We sweep over possible update schedules $u$ and assume the population has implemented critical social learning. All other simulation parameters are the same.


\subsection*{How Often Should an AI Update Its Model of the World?}
\label{supp-ai-individ-learn}

We then extend the base simulations to modify how the AI system learns at each step. The AI system either learns by snapping to the mean of the population, or ``pays'' a cost of individually learning. We vary the cost of individual learning and expected accuracy of the AI system individually learning. We assess the relative gains in the base scenario (when humans either individually or socially learn based on their reliance propensity) versus a population wherein people engage in critical social learning (i.e., they can switch strategies based on the success of social learning from the AI). We use the same parameters as in the previous sections.


\section*{Additional Details on Negative Feedback Environment Model}
\label{supp-feedback}

We extend the base simulations by adding an additional ``individual learning penalty'' property to each agent. The default value of this property is $\kappa=1$ in all previous simulations (i.e., social learning from the AI has no impact on an agent's ability to individually learn); here, we consider that each time an agent learns from the AI system, we multiply their penalty parameter by a scaling factor ($\kappa_j^{0}=1, \kappa_j^{t+1}=0.9\kappa_j^{t}$ for agent $j$ who learns from the AI). An agent's individual learning success probability is then the product of their parameter value and the base individual learning success probability in the environment ($p_{ij}^{OK}=(1-c_i)z_i\kappa_j$). All other parameters are held the same as in the base simulations.


\newpage







\end{document}